\documentclass[sigconf, nonacm]{acmart}
\usepackage{multirow}
\usepackage{multicol}
\usepackage{graphicx}
\usepackage{float}
\usepackage[table]{xcolor}
\usepackage{tikz}
\usetikzlibrary{arrows.meta, positioning}
\newcommand\vldbdoi{XX.XX/XXX.XX}
\newcommand\vldbpages{XXX-XXX}
\newcommand\vldbvolume{14}
\newcommand\vldbissue{1}
\newcommand\vldbyear{2020}
\newcommand\vldbauthors{\authors}
\newcommand\vldbtitle{\shorttitle} 
\newcommand\vldbavailabilityurl{https://github.com/nikagrwal/Benchmarking-KV-Cache-Optimizations-across-Task-Quality-and-System-Performance}
\newcommand\vldbpagestyle{plain} 

\begin{document}
\title{Benchmarking KV-Cache Optimizations across Task Quality and System Performance for Long-Context Serving [Experiment, Analysis \& Benchmark]}

\author{Nikita Agrawal}
\affiliation{%
  \institution{University of Bayreuth}
  \city{Bayreuth}
  \state{Germany}
  \postcode{95447}
}
\email{Nikita.Agrawal@uni-bayreuth.de}

\author{Ruben Mayer}
\affiliation{%
  \institution{University of Bayreuth}
  \city{Bayreuth}
  \state{Germany}
  \postcode{95447}
}
\email{Ruben.Mayer@uni-bayreuth.de}

\begin{abstract}

Large language model serving is increasingly limited by KV-cache growth under long-context workloads, yet existing KV-cache compression techniques are difficult to compare because they were evaluated on different models, tasks, budgets, and serving stacks. This paper presents a workload-aware benchmark of representative KV-cache optimization mechanisms spanning quantization, pruning, and merging, including KIVI, TurboQuant, SnapKV, and CaM, evaluated on LongBench-style multi-document QA, single-document QA, few-shot learning, and summarization workloads using Llama-3.1-8B-Instruct and Mistral-7B-Instruct-v0.3. The benchmark measures task quality, mean output throughput, mean time-to-first-token, and realized compression ratio across context-length buckets. The results show that the compression ratio alone is a poor predictor of end-to-end performance. KIVI4 provides the most stable quality across models, SnapKV delivers the strongest long-context throughput, and CaM yields large gains on selected QA workloads but exhibits substantial workload sensitivity in both quality and realized compression ratio. These findings motivate workload-aware selection of KV-cache mechanisms rather than one-size-fits-all compression and provide deployment guidance for long-context serving systems.

\end{abstract}

\maketitle

\pagestyle{\vldbpagestyle}
\begingroup\small\noindent\raggedright\textbf{PVLDB Reference Format:}\\
\vldbauthors. \vldbtitle. PVLDB, \vldbvolume(\vldbissue): \vldbpages, \vldbyear.\\
\href{https://doi.org/\vldbdoi}{doi:\vldbdoi}
\endgroup
\begingroup
\renewcommand\thefootnote{}\footnote{\noindent
This work is licensed under the Creative Commons BY-NC-ND 4.0 International License. Visit \url{https://creativecommons.org/licenses/by-nc-nd/4.0/} to view a copy of this license. For any use beyond those covered by this license, obtain permission by emailing \href{mailto:info@vldb.org}{info@vldb.org}. Copyright is held by the owner/author(s). Publication rights licensed to the VLDB Endowment. \\
\raggedright Proceedings of the VLDB Endowment, Vol. \vldbvolume, No. \vldbissue\ %
ISSN 2150-8097. \\
\href{https://doi.org/\vldbdoi}{doi:\vldbdoi} \\
}\addtocounter{footnote}{-1}\endgroup

\ifdefempty{\vldbavailabilityurl}{}{
\vspace{.3cm}
\begingroup\small\noindent\raggedright\textbf{PVLDB Artifact Availability:}\\
The source code, data, and/or other artifacts have been made available at \url{\vldbavailabilityurl}.
\endgroup
}

\section{Introduction}


The adoption of large language models (LLMs) has grown rapidly across a wide range of data-intensive applications, including  document summarization, multi-turn dialogue, and code analysis \cite{duan-etal-2024-botchat, 10.1145/3663741.3664785}. As these applications increasingly rely on long-context inputs, efficient LLM serving has emerged as a critical systems challenge. In particular, the Key–Value (KV) cache used in LLMs grows linearly with input length, leading to substantial memory and bandwidth overhead during inference \cite{10.1145/3778534.3778567}. This has motivated a surge of recent work on KV cache optimization techniques, such as quantization, pruning, and merging, to enable scalable long-context serving.

The data management community has recently begun to address these challenges by rethinking LLM serving stacks and system-level optimizations. Previous works have explored enhancing LLM serving by improving efficiency in inference pipelines, memory-aware scheduling, and hardware-conscious optimizations \cite{10.14778/3750601.3750703, 10.1145/3749168, 10.1145/3769778, 10.14778/3685800.3685838}. A growing research area addresses KV cache compression techniques to reduce memory footprint and improve decoding efficiency \cite{10.5555/3768039.3768067}. However, the proposed methods have only been evaluated in isolation or compared with few other techniques. The corresponding publications make a direct comparison difficult, as they use different models, datasets, compression budgets, and system configurations. Thus, it remains unclear how these methods compare among themselves, and which methods are most suitable for different workloads.

This lack of a unified and workload-aware evaluation is a key gap in the literature. In particular, existing studies often focus on either model quality or system efficiency in isolation, without jointly analyzing their trade-offs \cite{10.1145/3778534.3778567, yuan-etal-2024-kv}. Furthermore, there is limited understanding of how KV cache optimizations affect system metrics such as memory and bandwidth across various task types, input lengths, and model architectures. This slows the development of robust KV cache compression techniques and system-level optimization strategies.

In this paper, we present a comprehensive benchmark for KV cache optimization methods that jointly evaluates task quality and system performance under long-context workloads. We focus on representative inference-time techniques spanning three major paradigms: quantization, pruning, and merging. We employ widely used instruction-tuned models, Llama-3.1-8B-Instruct \cite{10.5555/3737916.3740097} and Mistral-7B-Instruct-v0.3 \cite{jiang2023mistral7b}, and cover both task-level accuracy and system-level efficiency.
Our benchmark is built on the LongBench suite \cite{bai2024longbench}, from which we select six datasets to evaluate task quality across four categories: multi-document QA, single-document QA, few-shot learning, and summarization. To assess system performance, we use three representative long-context datasets: NarrativeQA, GovReport, and Qasper. We measure key metrics, including time-to-first-token (TTFT), output throughput, and prefill KV cache memory footprint. This dual evaluation of quality and system performance enables us to capture the trade-offs between accuracy preservation and inference efficiency across different KV cache compression strategies.

Our main contributions are as follows:

\begin{itemize}
    \item We present a workload-aware benchmark that systematically evaluates KV cache compression methods across both task quality and system performance dimensions. This provides a holistic perspective on the methods and their associated trade-offs.
    \item We conduct a unified evaluation of representative methods from quantization (KIVI, TurboQuant), pruning and eviction (SnapKV) and merging (CaM) with consistent models, datasets, and experimental settings. This provides a fair comparison under realistic conditions.
    \item We study the tradeoffs between accuracy, throughput, latency, and memory footprint, and demonstrate that the compression ratio alone is not enough to evaluate the end-to-end performance of a KV compression approach. This insight has an immediate practical use and also guides future research on new KV cache optimizations.
\end{itemize}

The paper is organized as follows. Section~\ref{sec:background} introduces the necessary background of LLM serving and KV cache optimizations, including a detailed introduction of the techniques we use in our benchmark. In Section~\ref{sec:setup}, we provide a detailed account of the experimental setup, including workloads, setup and metrics. Based on that, we discuss the results in Section~\ref{sec:results}. From the results, we derive practical insights and lessons learned in Section~\ref{sec:lessons}. Finally, we discuss related work in Section~\ref{sec:related} before concluding the paper in Section~\ref{sec:conclusions}.
\section{Background}
\label{sec:background}

\begin{figure*}
    \centering\includegraphics[width=\linewidth, trim=125pt 200pt 125pt 20pt, clip]{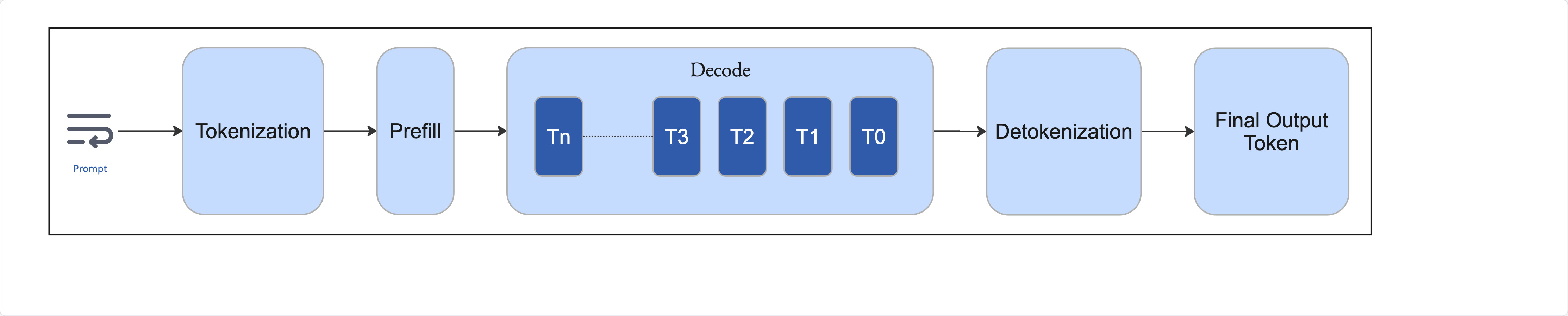}
    \caption{Prefill and decode during LLM inference. Token T0 is detokenized right after the prefill stage. After each token is generated in the decode step, the KV cache is updated. Detokenization happens after each decode step. Each token (T0, T1, T2, T3,....,Tn) is detokenized and output sequentially.}
    \label{fig:prefill-decode}
\end{figure*}

\subsection{LLM Inference Serving}
Large language model (LLM) inference serving refers to the system-level process of executing queries on pretrained models to generate outputs in real time. Fig. \ref{fig:prefill-decode} depicts a  typical serving pipeline, where an input prompt is first tokenized, i.e., converted from raw text into a sequence of discrete token IDs that can be processed by the model. The tokenized input is then processed during the prefill phase. In this phase, the entire input prompt is encoded, and the intermediate keys and values (KV) are computed and stored. This is followed by the \textit{decoding phase}, where tokens are generated autoregressively, one step at a time \cite{10.5555/3768039.3768067}. Each generated token is subsequently detokenized, i.e., mapped back from token IDs to human-readable text, enabling streaming output to the user.

\paragraph{\textbf{Prefill Phase.}}
Let $X \in \mathbb{R}^{b \times l_{\text{prompt}} \times d}$ denote the input tensor, where $b$ is the batch size, $l_{\text{prompt}}$ is the prompt length, and $d$ is the model hidden size. For simplicity, we omit the layer index. The key and value tensors are computed as
\begin{equation*}
X_K = X W_K, \quad X_V = X W_V
\end{equation*}
where $W_K, W_V \in \mathbb{R}^{d \times d}$ are the projection matrices for keys and values, respectively. Once computed, $X_K$ and $X_V$ are stored in the KV cache to facilitate efficient decoding \cite{10.5555/3692070.3693381}.

\paragraph{\textbf{Decoding Phase}} 
Let $t \in \mathbb{R}^{b \times 1 \times d}$ denote the current input token embedding. The corresponding key and value outputs are computed as $t_K = t W_K$ and $t_V = t W_V$, respectively. First, the KV cache is updated by appending the new entries:
\begin{equation*}
X_K \leftarrow \mathrm{Concat}(X_K, t_K), \quad
X_V \leftarrow \mathrm{Concat}(X_V, t_V).
\end{equation*}

Next, the attention output is computed as:
\begin{equation*}
\begin{aligned}
t_Q &= t W_Q, \\
A &= \mathrm{Softmax}(t_Q X_K^\top), \\
t_O &= A X_V,
\end{aligned}
\end{equation*}
where $W_Q$ denotes the query projection matrix. For simplicity, we omit the attention output projection layer and other components of the full inference pipeline \cite{10.5555/3692070.3693381}.

To avoid recomputing attention over all previous tokens at each decoding step, modern LLM systems maintain a \textit{KV cache}. For a sequence of length $N$, each Transformer layer stores key and value tensors corresponding to all past tokens. During decoding, the query vector of the current token attends to the cached keys and values, reducing the computational complexity from $O(N^2)$ to $O(N)$ per step \cite{liu2024cachegen}. This reuse of previously computed representations is critical for achieving low-latency inference.

However, the KV cache can introduce a major system bottleneck. Its memory footprint grows linearly with both sequence length and the number of layers, often dominating GPU memory usage in long-context scenarios \cite{10.1145/3749168}. For large models and long inputs, KV cache storage and access become the primary limiting factors for throughput and scalability. This has motivated many works on KV cache optimization techniques, including quantization, pruning, and merging, which aim to reduce memory usage while preserving model accuracy.

\subsection{Taxonomy of KV Cache Optimizations}

To compress the KV cache size, there have been proposed four principal methods. 

\paragraph{Quantization} The first principal method maintain all entries in the KV cache, but reduce the bit-length (precision) of the encoding of the stored keys and values. A common theme that runs among many KV quantization algorithm is mixed precision, where important tokens are kept at higher precision while heavily quantizing others. For example, ZipCache \cite{10.5555/3737916.3740097} and QAQ \cite{11375406} use attention-derived properties to identify important tokens and store them at higher precision. In ZipCache, a channel-wise token quantization is proposed to reduce parameter overhead, and a normalized attention score guides which tokens to preserve. Similarly, KIVI \cite{10.5555/3737916.3740097} finds that keys have a few large-magnitude channels while values do not. It thus applies per-channel quantization for keys and per-token quantization for values, enabling 2-bit KV storage with almost no accuracy loss. KVQuant \cite{10.5555/3737916.3737956} also exploits KV structure. It quantizes keys before rotary embeddings and isolates outliers per vector by keeping them in higher precision.

Another class uses vector transforms or codebooks to ease quantization. For example, CommVQ \cite{10.5555/3780338.3781795}  and PQCache \cite{11375406} apply vector quantization.  They split each KV vector into subvectors or additive code components and store compact indices instead of full float values. CommVQ \cite{10.5555/3780338.3781795} even designs its codebooks to commute with rotary embeddings, enabling as low as 1-bit precision with minimal loss.  PolarQuant \cite{wu2026polarquant} and TurboQuant \cite{zandieh2026turboquant} apply random rotations or coordinate changes: PolarQuant \cite{wu2026polarquant} transforms KV vectors to polar coordinates after a random rotation, yielding tightly distributed angles that can be quantized without extra scaling parameters. TurboQuant \cite{zandieh2026turboquant} shows that a random rotation makes each coordinate follow a concentrated Beta distribution, so simple optimal scalar quantizers can nearly achieve the theoretical best distortion.

Quantization approaches trade memory for extra computation or design complexity. Mixed-precision schemes require computing token importance using attention scores or norms, and per-channel methods need special grouping of data. Transform-based methods incur the cost of rotating or projecting vectors at decode time, and product quantization uses a cookbook for encoding/decoding. However, the extra computational cost provides orders-of-magnitude memory savings with only small drops in accuracy. Aggressive quantization increases compression but risks output quality. Methods that preserve outliers or adapt precision mitigate this at the cost of extra overhead \cite{10.5555/3737916.3737956, zandieh2026turboquant}.   

\paragraph{Pruning} The second principal method approaches the problem by compressing the KV cache along the token or structural dimension, either by discarding less important tokens or by sparsifying their representations. The core idea is to rank tokens by some importance signal and keep only a subset. Similar to quantization, many pruning methods use attention-derived scores to decide which tokens to retain. For example, SAGE-KV \cite{wang2025llms} computes attention after prefill and retains only the top-k tokens for inference. This exploits the ``attention sparsity” observation, usually only a few tokens have high influence. Similar approaches such as H2O \cite{10.5555/3666122.3667628} and SnapKV \cite{10.5555/3737916.3738638}, aggregate past attention to pick important tokens. Another simple yet effective approach is recency. Sliding-window schemes such as StreamingLLM \cite{xiao2024efficient} keep only  the most recent tokens plus a few early attention sinks to stabilize performance. These methods incur minimal overhead as no scoring of tokens is required, but can mistakenly drop older relevant context. Other signals such as L2 norm has been explored to guide eviction decisions \cite{devoto-etal-2024-simple}.

Beyond token level pruning, structured pruning removes fixed groups of tokens at once.  For example, PagedEviction operates on block ``pages” of the KV cache \cite{chitty-venkata-etal-2026-pagedeviction}. It evicts full blocks aligned to the memory layout, resulting in less fragmentation and overhead, but coarser granularity. The trade-off here is that aggressive pruning saves memory, but has the risk of losing useful context. Attention-based pruning is precise but adds computation; static policies are simpler but less adaptable. Empirically, hybrid-strategies often works best. In all cases, the main cost is potential accuracy loss. Evicted tokens cannot be recovered, so if importance is misjudged, model outputs degrade.

\paragraph{Merging} The third principal method aim to compress the KV cache by combining redundant or highly similar tokens into shared representations, rather than discarding them outright.  The core idea is to cluster redundant tokens and represent them with a shared state. For example, Adaptive KV \cite{ge2023model} clusters KV tokens by similarity and merges each cluster into one representative. Bolya et al. \cite{bolya2022tome} merges based on feature similarity on-the-fly during inference. Other work performs multi-level merging of tokens in a hierarchical manner, progressively compressing context while preserving high-level semantic structure \cite{song2024hierarchical}.

Merging incurs extra computation cost for calculating pairwise similarity or clustering similar tokens but can retain much of the original information compared to pruning. It avoids outright deletion of content by combining tokens that have less importance with the retained tokens, preserving their relevance in a compressed form \cite{ge2023model}. Few methods proposed by \cite{munkhdalai2024leavecontextbehindefficient} and \cite{kim2024compressed} progressively push past tokens into a fixed-size recurrent memory via a small updater network, so the KV cache acts like an RNN summary. This makes the KV cache quite complex and the merged states may lead to loss of fine grained details among the tokens. However, compared to pruning, merging yield smoother accuracy, since information is not fully lost.

\paragraph{Cross-Layer/Head Sharing and Low-Rank Methods} Quantization, pruning, and merging work within the token dimension of the KV cache. Another orthogonal approach compresses the KV cache across layers or attention heads, i.e., along the model's feature dimensions. They typically require model-level changes or retraining. A common theme is sharing KV projections. For example, Multi-Query Attention (MQA) \cite{shazeer2019fasttransformerdecodingwritehead} uses one shared key/value projection per layer instead of per head, whereas Grouped-Query Attention (GQA) \cite{ainslie2023gqa} uses a few projections shared among groups of heads.  MLKV \cite{zuhri-etal-2025-mlkv} extends this idea to multiple layers by sharing the same KV projections across layers, achieving up to 6× more compression beyond MQA. Similarly, YOCO \cite{sun2024you} introduces a two-stage transformer, which is a combination of a self-decoder and a cross-decoder, so that KV is computed globally once and reused by all layers. Other schemes such as \cite{10.5555/3737916.3742359} and KVSharer \cite{yang2024kvsharer} selectively merge KV states across layers based on similarity. These methods drastically cut KV size and bandwidth by design, often with some pre-training or architecture redesign. The downside is complexity. They require changes to the model architecture or training.

Another direction targets the feature dimension using low-rank approximations. Methods such as Palu \cite{chang2025palu} factorize attention into low-rank components and cache compressed representations, reconstructing full KV states on demand. LoRC \cite{zhang2025lorc} applies layer-wise low-rank projections to compress KV weights without retraining. By exploiting redundancy in hidden dimensions, these methods shrink KV storage, though they introduce approximation error and some extra compute for compression and reconstruction.

Overall, these methods structurally reduce KV size and are complementary to token-level techniques such as pruning or quantization. Their main drawback is integration complexity and potential accuracy loss, while their key advantage is a consistent reduction in memory usage and bandwidth, independent of input length.

\subsection{Benchmarked Techniques}

\textbf{Selection rationale for this study:} In our benchmarking study, we focus on drop-in replaceable KV cache compression techniques that do not require deep changes to the model architecture or even retraining. From the practical perspective of data management techniques, it is desirable that a data system architect or administrator can easily integrate the techniques into existing infrastructure. Following this rationale, we selected representative techniques from the three categories of quantization (KIVI, Turboquant), pruning (SnapKV), and merging (CaM). We discuss the details of these techniques in the following.

\paragraph{KIVI} The mathematical core of KIVI \cite{10.5555/3692070.3693381} is built upon the observation that the key cache ($K$) and value cache ($V$) exhibit different outlier structures. In the key cache, outliers are concentrated in specific channels, meaning certain dimensions across all tokens have much higher magnitudes. Conversely, the value cache contains outliers that are token-specific, where certain entire tokens have higher magnitudes across all channels. KIVI uses an asymmetric quantization scheme to map these high-precision floating-point values into a discrete $B$-bit integer space.

For any given tensor $X$, the quantization process is defined by finding a scaling factor $s_X$ and a zero-point $z_X$. The quantization function $Q(X)$ maps the input to the nearest integer in the range $[0, 2^B - 1]$ using the following formulas:

\begin{equation*}
\begin{aligned}
z_X = \min(X), \\
s_X = \frac{\max(X) - \min(X)}{2^B - 1}, \\
Q(X) = \left\lfloor \frac{X - z_X}{s_X} \right\rceil
\end{aligned}
\vspace{0.7em}
\end{equation*}


KIVI adopts different quantization axes for keys and values to align with their distinct outlier patterns. It applies \textit{per-channel quantization} to keys (column-wise) so that large values concentrated in certain dimensions do not distort the scaling for the entire tensor. For values, it applies \textit{per-token quantization} (row-wise) so that tokens with unusually large magnitudes are handled locally, preventing them from degrading the precision of other tokens.

During the inference phase, the B-bit quantization integers must be converted back to the original precision for the model to perform the attention operation. This process can be expressed as

\begin{equation*}
X' = Q(X) \cdot s_X + z_X
\end{equation*}

where $\lfloor \cdot \rceil$ denotes the rounding operator to the nearest integer. 

To further stabilize performance, KIVI employs a streaming strategy involving a residual cache. New tokens are kept in full precision because recent tokens often contribute most significantly to the attention output, and their distributions have not yet stabilized. Once the residual cache reaches a pre-defined threshold, the oldest tokens in the residual are quantized using the per-channel/per-token logic and appended to the compressed 2-bit KV cache. This hybrid precision approach ensures that the ``local'' context remains highly accurate while the ``distant'' context is efficiently compressed.

\paragraph{TurboQuant} TurboQuant \cite{zandieh2026turboquant} approaches transforms KV cache vectors into a distribution that is amendable to efficient quantization while preserving attention-relevant inner products. Given an input vector $x \in \mathbb{R}^d$, TurboQuant first applies a randomized orthogonal transformation using a matrix $Q \in \mathbb{R}^{d \times d}$, producing $y = Qx$. This transformation preserves the Euclidean norm while redistributing the vector’s energy uniformly across dimensions. In high-dimensional settings, the coordinates $y_i$ approximate a Gaussian distribution, which enables the use of simple, data-independent scalar quantizers.

For a given vector $y$, TurboQuant performs coordinate-wise $b$-bit quantization. Let $Q_s(\cdot)$ denote the scalar quantizer; the quantized vector is defined as
\begin{equation*}
\hat{y} = [Q_s(y_1), Q_s(y_2), \dots, Q_s(y_d)]^\top,
\end{equation*}
with total distortion
\begin{equation*}
\mathrm{MSE}(y, \hat{y}) = \sum_{i=1}^{d} \mathbb{E}[(y_i - \hat{y}_i)^2].
\end{equation*}
By leveraging the Gaussian-like distribution induced by the rotation, this approach achieves near-optimal rate--distortion performance using simple scalar quantization.

To mitigate the contraction bias introduced by MSE-optimal quantization, TurboQuant incorporates a residual correction mechanism. After quantization, the residual vector $r = y - \hat{y}$ is computed and encoded using a 1-bit Quantized Johnson--Lindenstrauss (QJL) transform, storing only the sign information:
\begin{equation*}
s = \mathrm{sign}(r).
\end{equation*}
During inference, the quantized vector $\hat{y}$ and residual signs $s$ are combined to produce an unbiased estimate of inner products, ensuring that attention scores remain accurate despite aggressive compression.

This design enables TurboQuant to compress KV cache vectors to very low bit rates while maintaining high fidelity in both reconstruction error and attention computation.

\paragraph{SnapKV} SnapKV \cite{10.5555/3737916.3738638} leverages the intrinsic sparsity and consistency of attention patterns in LLMs to compress the KV cache. The method identifies salient historical tokens by observing the attention distribution at the end of a prompt and evicting non-essential entries to maintain a constant-sized cache. This process is governed by a heuristic selection mechanism that utilizes pooled attention scores to determine which KV pairs are critical for future token generation.

For a prompt of length $N$, SnapKV computes attention weights using an observation window of queries $Q_{\text{obs}}$ attending over prefix keys $K_{\text{prefix}}$:
\begin{equation*}
A = \mathrm{Softmax}\left( \frac{Q_{\text{obs}} K_{\text{prefix}}^\top}{\sqrt{d}} \right).
\end{equation*}

To estimate token importance, the attention weights are aggregated across the observation window and smoothed using a 1D average pooling operation:
\begin{equation*}
S = \mathrm{AvgPool}(A, \text{kernel}=s).
\end{equation*}
The most salient tokens are then selected via a Top-$k$ operation under a fixed cache budget $C$:
\begin{equation*}
I = \mathrm{Top}\text{-}k(S, k = C - L_{\text{recent}}).
\end{equation*}

The final compressed KV cache consists of the union of selected salient tokens and a recent token window:
\begin{equation*}
KV_{\text{compressed}} = \{ KV_i \mid i \in I \} \cup \{ KV_j \mid j > N - L_{\text{recent}} \}.
\end{equation*}

By keeping both $C$ and $L_{\text{recent}}$ fixed, SnapKV preserves long-range dependencies through salient token retention while maintaining local coherence via recent tokens, enabling efficient long-context inference without retraining.

\paragraph{CaM} Cache Merging (CaM) departs from conventional KV cache pruning methods by avoiding hard eviction and instead redistributing the contribution of removed tokens into retained ones. The key idea is to preserve the attention output by merging values of evicted tokens into nearby tokens, thereby reducing the bias introduced by removing low-probability but non-negligible contributions.

In a standard attention layer, the output for a query is given by
\begin{equation*}
O = \sum_{i=1}^{n} \alpha_i V_i,
\end{equation*}
where $\alpha_i$ are attention weights and $V_i$ are value vectors. When a token $k$ is removed, CaM merges its contribution into a retained token $j$ by updating
\begin{equation*}
V_j' = V_j + \frac{\alpha_k}{\alpha_j} V_k,
\end{equation*}
which preserves the attention output since
\begin{equation*}
\alpha_j V_j' = \alpha_j V_j + \alpha_k V_k.
\end{equation*}
This formulation shows that, given exact attention ratios, merging can be lossless.

Since future attention weights are unknown during inference, CaM approximates this process using \textit{even merging}, where an evicted token $V_i$ is distributed across a local window of $m$ retained tokens:
\begin{equation*}
V_j' = V_j + \frac{1}{m} V_i.
\end{equation*}
This approximation assumes locally similar attention magnitudes and reduces variance compared to single-point merging.

To further improve robustness, CaM employs an adaptive merging strategy based on cumulative attention scores $\bar{A}_i$, which provide a stable estimate of token importance. The decision to merge a token is modeled as
\begin{equation*}
M_i \sim \mathrm{Bernoulli}\left( \mathrm{clamp} \left( \frac{\bar{A}_i}{\frac{1}{m} \sum_{j \in \text{window}} \bar{A}_j}, 0, 1 \right) \right).
\end{equation*}
Tokens are merged when their relative importance is sufficiently high; otherwise, they are evicted.

Overall, CaM compresses the KV cache by converting discrete token removal into a continuous redistribution of value representations. This design preserves attention outputs more faithfully than standard pruning while maintaining a bounded memory footprint.

\section{Experimental Setup}
\label{sec:setup}

\begin{table}[t]
\centering

\renewcommand{\arraystretch}{1.0}
\setlength{\tabcolsep}{3pt}
\begin{tabular}{|l|c|c|}
\hline
\textbf{Dataset} & \textbf{Min-Max Tokens} & \textbf{Workload} \\
\hline
HotpotQA         & 2K-8K  & Multi-Doc QA \\
2WikiMQA         & 2K-8K  & Multi-Doc QA \\
Qasper           & 1K-20K & Single-Doc QA \\
MultiFieldQA\_en & 1K-16K & Single-Doc QA \\
TriviaQA         & 2K-8K  & Few-shot Learning  \\
Multi-news       & 4K-32K & Summarization \\
NarrativeQA      & 8K-64K & Single-Doc QA \\
GovReport        & 2K-64K & Summarization \\
\hline
\end{tabular}
\caption{Datasets, token ranges, and workload types.}
\label{tab:datasets}
\end{table}

\paragraph{Workloads} We evaluate KV cache compression methods using the LongBench benchmark \cite{bai2024longbench}, which consists of many long-context tasks. Table \ref{tab:datasets} shows the datasets we use along with the minimum and maximum context length test examples it consist of and the workload type. We consider four varying categories: (1) multi-document QA, which needs to extract
and combine information from several documents
to obtain the answer, (2) single-document QA, which tests the long
context understanding ability with longer documents, (3) few-shot learning, which is a practical
setting requiring long-context understanding over
provided examples, and (4) summarization, which
requires a global understanding of the whole context in this work \cite{yuan-etal-2024-kv}.
For task quality evaluation, we use six representative datasets: HotpotQA and 2WikiMQA for multi-document QA, Qasper and MultiFieldQA\_en for single-document QA, TriviaQA for few-shot learning, and MultiNews for summarization. These datasets span a wide range of context lengths, reasoning complexity, and dependency patterns, from multi-hop retrieval across documents to long-form generation tasks. For system efficiency evaluation, we focus on NarrativeQA, Qasper, and GovReport, as they contain substantially longer contexts, making them better suited to capturing realistic KV cache behavior and more representative system-level performance under long-context workloads. 
This enables us to systematically study the impact of KV cache compression on task quality and system efficiency for various long context tasks. 

\paragraph{Compared Methods} 
We select a small but representative set of KV cache compression methods to capture the core design trade-offs relevant for long-context serving systems. Rather than exhaustively evaluating all prior work, we focus on widely adopted, training-free methods that can be directly applied at inference time and reflect distinct system-level behaviors.

For quantization, we evaluate KIVI \cite{10.5555/3692070.3693381} and TurboQuant \cite{zandieh2026turboquant}, which represent two complementary techniques. KIVI is a data-dependent approach that explicitly models the outlier structure in keys and values via asymmetric quantization, achieving consistent accuracy despite aggressive compression. On the other hand, TurboQuant is a data-oblivious, rotation-based approach with strong theoretical guarantees, allowing efficient low-bit compression that behaves distinctly under varying workloads. For pruning and eviction, we include SnapKV \cite{10.5555/3737916.3738638}, a state-of-the-art training-free approach that leverages consistent attention patterns to retain salient tokens under a fixed cache budget. This makes it representative of practical token-level sparsification strategies used in long-context inference. For merging-based compression, we evaluate CaM \cite{pmlr-v235-zhang24n}, which departs from hard eviction by redistributing evicted token contributions to preserve attention outputs. This provides a fundamentally different trade-off, prioritizing output integrity over strict sparsity.

These KV cache compression methods can be easily deployed in real systems. It enables us to systematically analyze how different KV cache optimization strategies impact both task quality and end-to-end system performance under diverse workloads.

\paragraph{LLMs} 
We conduct experiments on two widely used instruction-tuned models: Llama-3.1-8B-Instruct \cite{grattafiori2024llama3herdmodels} and Mistral-7B-Instruct-v0.3 \cite{jiang2023mistral7b}. Both models use Grouped-Query Attention (GQA). The differentiator is that Mistral-7B specifically utilizes Sliding Window Attention (SWA) and a rolling buffer cache for extreme efficiency, whereas Llama-3.1-8B focuses on maximizing reasoning performance, utilizing a much larger 128k context window compared to 32k in Mistral-7B.

These architectural differences lead to different attention behaviors and KV cache usage patterns, providing a diversity of models for evaluating compression strategies. Using both models allows us to determine whether the observed performance trends are consistent across LLM families or sensitive to model-specific characteristics.

\paragraph{Setup} 
All methods were evaluated using the same models and datasets. Models were loaded in FP16 with FlashAttention2 \cite{dao2024flashattention} enabled, and prompts were formatted using the corresponding chat template for each instruction-tuned model. We used the LongBench prompt templates and dataset-specific generation budgets~\cite{bai2024longbench}. 

For KIVI \cite{10.5555/3692070.3693381}, we use the official KIVI-style asymmetric KV-cache quantization implementation integrated into the Llama and Mistral attention modules. We evaluate both 2-bit and 4-bit KV-cache variants. In both cases, keys and values are quantized to the same bit width, with group size for group-wise quantization and residual length for preserving the latest tokens in full precision as 32. The KIVI's FP16 model where both keys and values are stored in full precision without any quantization is used as the full-cache baseline.

For TurboQuant \cite{zandieh2026turboquant}, we used the Transformers 4.45-compatible TurboQuant cache implementation by Omar Hory \cite{hory2026turboquant}. We evaluated 3-bit and 4-bit KV-cache quantization. The 3-bit configuration used outlier-aware quantization: for a head dimension of 128, 32 outlier channels were stored at 4 bits, giving an effective precision of 3.25 bits per value. The 4-bit setting used uniform 4-bit quantization without outlier channels. The reported runs used the unpacked cache representation, so the reported memory usage may be slightly higher than with additional bit-packing optimizations. 

For SnapKV, we used the SnapKVPress implementation from KVPress \cite{devoto2026expected}. The cache was compressed just after prefill using SnapKV’s attention-based token selection \cite{10.5555/3737916.3738638}. We set the compression ratio to 0.75. In SnapKV, window size denotes the number of most recent query tokens used to estimate the importance of earlier KV entries, while kernel size denotes the width of the average-pooling filter used to smooth the resulting attention-based importance scores. We set the window size to 32 and the kernel size to 7.

For CaM \cite{pmlr-v235-zhang24n}, we evaluated the CAMPress wrapper with SnapKV as the base compressor provided by KVPress \cite{devoto2026expected}. We used CAMPress with a compression interval of 32, which means that CaM applies one merge-and-prune compression pass after every 32 decoding steps. After each compression pass, the KV cache is reduced to a pre-defined number of retained tokens per layer, known as the target cache size. We set it to 1024. The hidden-state buffer size is set to 64, which provides longer recent decoding context available for scoring and compressing. A merge budget of 32 helps redistribute the token value across up to 32 subsequent tokens instead of being simply dropped. The underlying SnapKV base press uses a window size of 16 and a kernel size of 7. We use a smaller SnapKV window inside CaM because CaM performs periodic decoding-time compression using recent buffered hidden states and cumulative attention, so a shorter observation window emphasizes the most recent local decoding context and keeps the base scorer aligned with interval-based compression. Thus, CaM performs periodic cache merging during generation rather than only applying a one-shot prefill compression.

\begin{table*}[t]
\centering
\renewcommand{\arraystretch}{1.2}
\setlength{\tabcolsep}{7pt}
\begin{tabular}{|c|c|cc|cc|c|c|}
\hline
 &  & \multicolumn{2}{c|}{Multi-Doc QA} & \multicolumn{2}{c|}{Single-Doc QA} & Few-Shot Learning & Summarization \\ 
\cline{3-4} \cline{5-6} \cline{7-7} \cline{8-8}
 & \multicolumn{1}{c|}{Models} &   HotpotQA &  2WikiMQA & Qasper & MultifieldQA\_en & Trivia\_QA & Multi-news \\
\hline

\multirow{7}{*}{\rotatebox{90}{\makebox[0pt][c]{\scriptsize Llama-3.1-8b-Instruct}}}

& All KV & 55.72 & 44.27 & 45.76 & 53.92 & 91.65 & \textbf{27.15} \\
\cline{2-8}
&  KIVI2        & 54.42 & 42.94 & 44.07 & 54.98 & \textbf{92.56} & \textit{26.99} \\ \cline{2-8}
&  KIVI4        & 55.99 & 46.63 & 45.42 & 54.74 & 91.66 & 26.95 \\ \cline{2-8}
&  TurboQuant3  & 53.19 & 41.57 & 46.60 & 51.46 & 84.84 & 16.39* \\ \cline{2-8}
&  TurboQuant4  & 56.10 & 45.07 & 46.03 & 53.31 & 90.13 & 16.16* \\ \cline{2-8}
&  SnapKV-0.75   & \textit{59.2} & \textit{49.41} & \textit{46.37} & \textit{56.06} & \textit{92.04} & 23.04 \\ \cline{2-8}
&  CaM  & \textbf{59.8} & \textbf{51.33} & \textbf{47.32} & \textbf{56.42} & 91.71 & 17.56 \\ \hline\hline

\multirow{7}{*}{\rotatebox{90}{\makebox[0pt][c]{\scriptsize Mistral-7b-Instruct-v0.3}}}
& All KV        & \textbf{51.54} & 36.16 & 38.27 & 50.10 & \textbf{88.59} & 26.42 \\ \cline{2-8}
&  KIVI2        & 48.46 & 38.34 & 39.89 & \textbf{54.12} & 87.82 & \textit{27.04} \\ \cline{2-8}
&  KIVI4        & 48.98 & \textbf{39.11} & \textbf{41.10} & \textit{53.56} & \textit{88.54} & \textbf{27.68} \\ \cline{2-8}
&  Turboquant3  & \textit{49.45} & 34.31 & 37.16 & 50.21 & 88.05 & 15.86* \\ \cline{2-8}
&  Turboquant4  & 49.43 & 33.94 & 40.39 & 50.65 & 88.09 & 16.01* \\ \cline{2-8}
&  SnapKV-0.75   & 48.43 & \textit{38.46} & 36.44  & 51.23 & 86.31 & 24.25 \\ \cline{2-8} 
&  CaM   & 48.57 & 37.59 & \textit{40.55} & 52.4 & 85.37 & 16.74 \\ \hline

\end{tabular}
\vspace{1em}
\caption{Accuracy results on LongBench tasks for Llama-3.1-8B-Instruct and Mistral-7B-Instruct-v0.3 under different KV compression methods. Bold and italic values denote the best and second-best performance per column, highlighting the trade-offs between compression efficiency and task-specific accuracy. \textit{Note: *}Summarization task using TurboQuant was only ran once as the run takes 3-4 days to complete.}
\label{accuracy_over_tasks}
\end{table*}

We conduct our experiments on a NVIDIA A100 GPU with 40GB of memory. As a result, we limit our evaluation to models in the 7-8B parameter range, which represents a practical balance between model capability and
feasible long-context inference under KV cache compression.
The main trends we observe are expected to generalize to larger models and distributed settings. KV cache size and memory bandwidth costs grow with model size and context length, so KV compression carries the same importance for larger models \cite{10.5555/3692070.3693381}. The workload-dependent patterns we observe are driven by task structure and are expected to persist. However, exact memory usage, throughput, and latency will depend on system-level optimizations, so quantitative gains may differ in larger deployments.

\paragraph{Metrics} 
We evaluate KV cache compression methods along two dimensions: \textit{task quality} and \textit{system efficiency}.

We report both per-task performance and category-level averages to capture method behavior across heterogeneous workloads. In doing so, we rely on accuracy metrics provided by LongBench for each dataset \cite{bai2024longbench}. For the question answering tasks HotpotQA, 2WikiMQA, Qasper, and MultiFieldQA\_en, we use the F1 score, which measures the overlap between predicted and ground-truth answers. For  TriviaQA, we use the exact match (EM), which checks whether the model’s predicted answer matches the ground-truth answer exactly, after applying basic normalization. For summarization datasets such as MultiNews, we report ROUGE scores, which evaluate the quality of generated summaries based on n-gram overlap with reference summaries. These metrics reflect the model’s ability to preserve semantic correctness and coherence under KV cache compression. We report both per-dataset performance and category-level averages.

We focus our system-level evaluation on three representative datasets: NarrativeQA, GovReport, and Qasper, as they have longer context lengths. We measure efficiency metrics such as time-to-first-token (TTFT), output throughput, and prefill KV memory. 

\textit{Time-to-First-Token} measures the time elapsed from when a user submits a prompt to when the model generates its very first output token. It captures the total latency of the prefill stage, including KV cache construction and initial model overhead. It is heavily influenced by input prompt length, KV cache compression strategies, and includes the computation time required to generate the very first output token. Lower TTFT indicates faster perceived responsiveness in real-time applications such as chatbots, voice AI, and interactive streaming because it reduces the initial wait time before the user sees the first piece of output.

\textit{Output Throughput} 
is measured as the average number of tokens generated per second during inference. This metric reflects the efficiency of the decoding stage and is directly impacted by the KV cache compression used. Higher throughput indicates better system performance.

\textit{Prefill KV memory} measures how much VRAM is consumed to hold the context during that initial KV processing stage, which is a major factor in overall peak memory. This metric captures the effectiveness of each compression method in reducing memory usage. Lower KV cache memory enables longer context lengths and improved scalability under given hardware constraints.

\section{Results}
\label{sec:results}

\begin{table*}[t]
\centering
\renewcommand{\arraystretch}{1.2}
\setlength{\tabcolsep}{5pt}
\begin{tabular}{|c|c|c|ccccccc|}
\hline
 &  &  & \multicolumn{7}{c|}{Time to first token (in ms)} \\ \cline{4-10}
 &  & \textbf{Context} & \textbf{All KV}& \textbf{KIVI2} & \textbf{KIVI4} & \textbf{TurboQuant3} & \textbf{TurboQuant4} & \textbf{SnapKV-0.75} & \textbf{CaM} \\
\hline

\multirow{8}{*}{\rotatebox{90}{\makebox[0pt][c]{\scriptsize Llama-3.1-8b-Instruct}}}
& \multirow{2}{*}{ NarrativeQA}
 &  4-8K & 724.89 & 782.14 & 777.80 & 882.29 & 812.66 & 763.85 & 744.18 \\ \cline{3-10}
& &  8K+ & 3873.34 & 4048.71 & 4039.81 & 4397.15 & 4273.93 & 3982.49 & 3922.54 \\ \cline{2-10}

& \multirow{3}{*}{ GovReport}
&  0-4K & 287.42 & 334.59 & 332.31 & - & - & 310.93 & 306.06 \\ \cline{3-10}
& &  4-8K & 552.74 & 605.68 & 603.34 & - & - & 581.63 & 572.97 \\ \cline{3-10}
& &  8K+ & 1402.18 & 1485.98 & 1482.85 & - & - & 1449.56 & 1430.83 \\ \cline{2-10}

& \multirow{3}{*}{ Qasper}
&  0-4K & 289.62 & 333.31 & 330.20 & 405.08 & 324.25 & 310.62 & 306.25 \\ \cline{3-10}
& &  4-8K & 496.82 & 544.73 & 541.85 & 637.29 & 558.41 & 522.12 & 514.27 \\ \cline{3-10}
& &  8K+ & 1124.66 & 1194.20 & 1190.24 & 1351.43 & 1260.67 & 1163.29 & 1148.08 \\ \hline

\multirow{7}{*}{\rotatebox{90}{\makebox[0pt][c]{\scriptsize Mistral-7b-Instruct-v0.3}}}
& \multirow{1}{*}{NarrativeQA}
&  8K+ & 4932.28 & 5013.10 & 5042.05 & 4757.37 & 4172.87 & 4957.70 & 4905.75 \\ \cline{2-10}

& \multirow{3}{*}{GovReport}
&  0-4K & 308.67 & 346.50 & 348.43 & - & - & 317.84 & 314.13 \\ \cline{3-10}
& &  4-8K & 556.53 & 592.91 & 595.35 & - & - & 563.84 & 555.55 \\ \cline{3-10}
& &  8K+ & 1504.87 & 1547.04 & 1556.57 & - & - & 1508.24 & 1486.31 \\ \cline{2-10}

& \multirow{3}{*}{Qasper}
&  0-4K & 311.67 & 346.38 & 346.39 & 497.89 & 338.68 & 319.88 & 316.52 \\ \cline{3-10}
& &  4-8K & 532.64 & 567.66 & 568.23 & 778.63 & 582.58 & 541.21 & 534.14\\ \cline{3-10}
& &  8K+ & 1138.46 & 1176.59 & 1182.53 & 1513.77 & 1249.65 & 1143.80 & 1133.55 \\ \hline

\end{tabular}
\vspace{1em}
\caption{Time to first token of different KV caching methods across 3 datasets (NarrativeQA, GovReport, Qasper) and varying context length ranges for Llama-3.1-8B-Instruct and Mistral-7B-Instruct-v0.3. \\ \textit{Note:} We do not present results for Gov\_Report using TurboQuant because it takes up to 3 days for one single run.}
\label{ttft}
\end{table*}

Our results are organized into two sections: \textit{Task Quality} and \textit{System Efficiency}.

\subsection{Task Quality}

We first discuss general observations and trends across all tasks before discussing the detailed results for each workload in separate paragraphs. Table~\ref{accuracy_over_tasks} shows all accuracy results.

\paragraph{General Observations}
We observe clear differences between KV-cache compression paradigms across all workloads and models. Overall, moderate quantization methods, especially KIVI4, achieve the most stable accuracy across workloads, while pruning using SnapKV and merging using CaM often yield higher peak accuracy across the QA workload, but the accuracy is variable across other tasks. This can be attributed to the fact that, despite aggressive compression, KIVI handles key/value outliers in an efficient manner. Aggressive quantization, such as in TurboQuant, leads to a drop in quality, especially for generation-heavy workloads such as summarization. This indicates that, although rotation-based quantization is theoretically efficient, it introduces reconstruction errors that negatively affect tasks requiring global context understanding.   



\paragraph{Multi-Document QA} For Multi-Document QA, CaM and SnapKV achieve good performance across both models. On Llama-3.1-8B, both CaM and SnapKV even outperform the All-KV baseline. This can be explained by the nature of the Multi-Doc QA task, which requires the retrieval and aggregation of sparse but highly relevant information across documents. Pruning preserves important tokens based on attention scores, ensuring that important cross-document context is retained, and merging redistributes information instead of discarding it, which reduces information loss compared to pruning. Heavy quantization with KIVI2 and TurboQuant3 shows slight performance degradation. This is likely because precision loss in attention scores affects multi-hop reasoning, where small numerical differences can propagate across reasoning chains. 

\paragraph{Single-Document QA} In Single-Document QA tasks, results are more balanced, especially with the Llama model. Similar to Multi-Document QA, CaM achieves the best performance across all the compression methods with Llama, while KIVI4 performs consistently well across both models. 

The good performance of CaM indicates that merging benefits tasks that require holistic document understanding, as it preserves broader context. KIVI4 achieves the best Qasper accuracy score and near-best MultifieldQA scores for Mistral-7b-Instruct-v0.3. The relatively weaker performance of SnapKV on Mistral suggests that aggressive token selection may remove context needed for detailed comprehension, especially in long single documents.

\paragraph{Few-Shot Learning} In few-shot learning, quantization methods perform surprisingly well, with KIVI2 achieving the highest score. This suggests that few-shot learning tasks are less sensitive to fine-grained KV precision, as they rely more on pattern recognition from examples rather than on exact long-range dependencies. Few-shot learning workload is robust to compression as long as recent tokens, which here are few-shot examples, are preserved. KIVI explicitly ensures this via its residual cache mechanism. 

In contrast, aggressive TurboQuant compression using the Llama-3.1-8b-Instruct model shows a noticeable degradation in quality of almost 7\%. This indicates that aggressive low-bit quantization may still harm the pattern extraction required for few-shot learning tasks. 

SnapKV and CaM remain competitive but do not outperform KIVI. This can likely be attributed to the fact that token selection and eviction or merging provide less benefit when most few-shot examples are already highly relevant.

\paragraph{Summarization} Summarization workload exhibits the largest largest performance drop across compression methods, especially for TurboQuant and CaM. 

Quantization using KIVI remains closest to baseline All-KV quality performance, mainly because KIVI retains all tokens, only reducing precision rather than changing the token structure. TurboQuant underperforms in summarization because its uniform quantization introduces small errors across all tokens, which accumulate and disrupt the global context coherence required for high-quality summaries. SnapKV shows a moderate degradation in quality of approximately 15\% and 8\% using Llama-3.1-8b-Instruct and Mistral-7b-Instruct-v0.3 respectively. This reflects that pruning might remove information that may still be important for coherent summaries. Whereas merging, especially using CaM, introduces approximation errors when combining tokens, which results in a quality degradation of approximately 35-36\%  using both models.

Thus, summarization is most sensitive to structural modifications of the KV cache, confirming that compression methods that preserve full context are preferable.

\subsection{System Efficiency}

\begin{table*}[t]
\centering
\renewcommand{\arraystretch}{1.2}
\setlength{\tabcolsep}{4pt}
\begin{tabular}{|c|c|c|c|c|cccccc|}
\hline
 &  &  &  &  & \multicolumn{6}{c|}{Compression Rate} \\ \cline{6-11}
 &  & \textbf{Context} & \textbf{Min\_tokens} & \textbf{Max\_Tokens} & \textbf{KIVI2} & \textbf{KIVI4} & \textbf{TurboQuant3} & \textbf{TurboQuant4} & \textbf{SnapKV-0.75} & \textbf{CaM} \\
\hline

\multirow{8}{*}{\rotatebox{90}{\makebox[0pt][c]{\scriptsize Llama-3.1-8b-Instruct}}}
& \multirow{2}{*}{ NarrativeQA}
 &  4-8K &  7964 &  7964 & 5.25 & 3.17 & 4.27 & 3.76 & 4.00 & 1.00 \\ \cline{3-11}
& &  8K+ &  8962 &  65271 & 5.32 & 3.19 & 4.27 & 3.76 & 4.00 & 1.00 \\ \cline{2-11}

& \multirow{3}{*}{ GovReport}
&  0-4K &  2020 &  3919 & 5.16 & 3.15 & - & - & 4.00 & 3.09 \\ \cline{3-11}
& &  4-8K &  4138 &  7979 & 5.24 & 3.17 & - & - & 4.00 & 6.04 \\ \cline{3-11}
& &  8K+ &  8118 &  51393 & 5.29 & 3.18 & - & - & 4.00 & 13.82 \\ \cline{2-11}

& \multirow{3}{*}{ Qasper}
&  0-4K &  1847 &  3934 & 5.16 & 3.15 & 4.27 & 3.76 & 4.00 & 1.26 \\ \cline{3-11}
& &  4-8K &  4044 &  7874 & 5.23 & 3.17 & 4.27 & 3.76 & 4.00 & 1.17 \\ \cline{3-11}
& &  8K+ &  8027 &  21111 & 5.28 & 3.18 & 4.27 & 3.76 & 4.00 & 1.22 \\ \hline

\multirow{7}{*}{\rotatebox{90}{\makebox[0pt][c]{\scriptsize Mistral-7b-Instruct-v0.3}}}
& \multirow{1}{*}{NarrativeQA}
&  8K+ &  9570 &  81496 & 5.31 & 3.19 & 5.02 & 4.43 & 4.00 & 1.00 \\ \cline{2-11}

& \multirow{3}{*}{GovReport}
&  0-4K &  2204 &  3988 & 5.16 & 3.15 & - & - & 4.00 & 3.21 \\ \cline{3-11}
& &  4-8K &  4140 &  7989 & 5.32 & 3.17 & - & - & 4.00 & 5.86 \\ \cline{3-11}
& &  8K+ &  8071 &  58334 & 5.30 & 3.18 & - & - & 4.00 & 14.20 \\ \cline{2-11}

& \multirow{3}{*}{Qasper}
&  0-4K &  2091 &  3986 & 5.17 & 3.15 & 4.27 & 3.76 & 4.00 & 1.09 \\ \cline{3-11}
& &  4-8K &  4002 &  7841 & 5.24 & 3.17 & 4.27 & 3.76 & 4.00 & 1.07 \\ \cline{3-11}
& &  8K+ &  8024 &  24121 & 5.28 & 3.18 & 4.27 & 3.76 & 4.00 & 1.13 \\ \hline

\end{tabular}
\vspace{1em}
\caption{Compression rates of different KV caching methods across 3 datasets (NarrativeQA, GovReport, Qasper) and varying context length ranges for Llama-3.1-8B-Instruct and Mistral-7B-Instruct-v0.3. We report compression ratios alongside minimum and maximum token lengths for each method, illustrating how compression efficiency varies with sequence length and task. \\ \textit{Note:} We do not present results for Gov\_Report using TurboQuant because it takes upto 3 days for one single run.}
\label{compression_rates}
\end{table*}

\begin{figure}
    \centering
    \includegraphics[width=\linewidth]{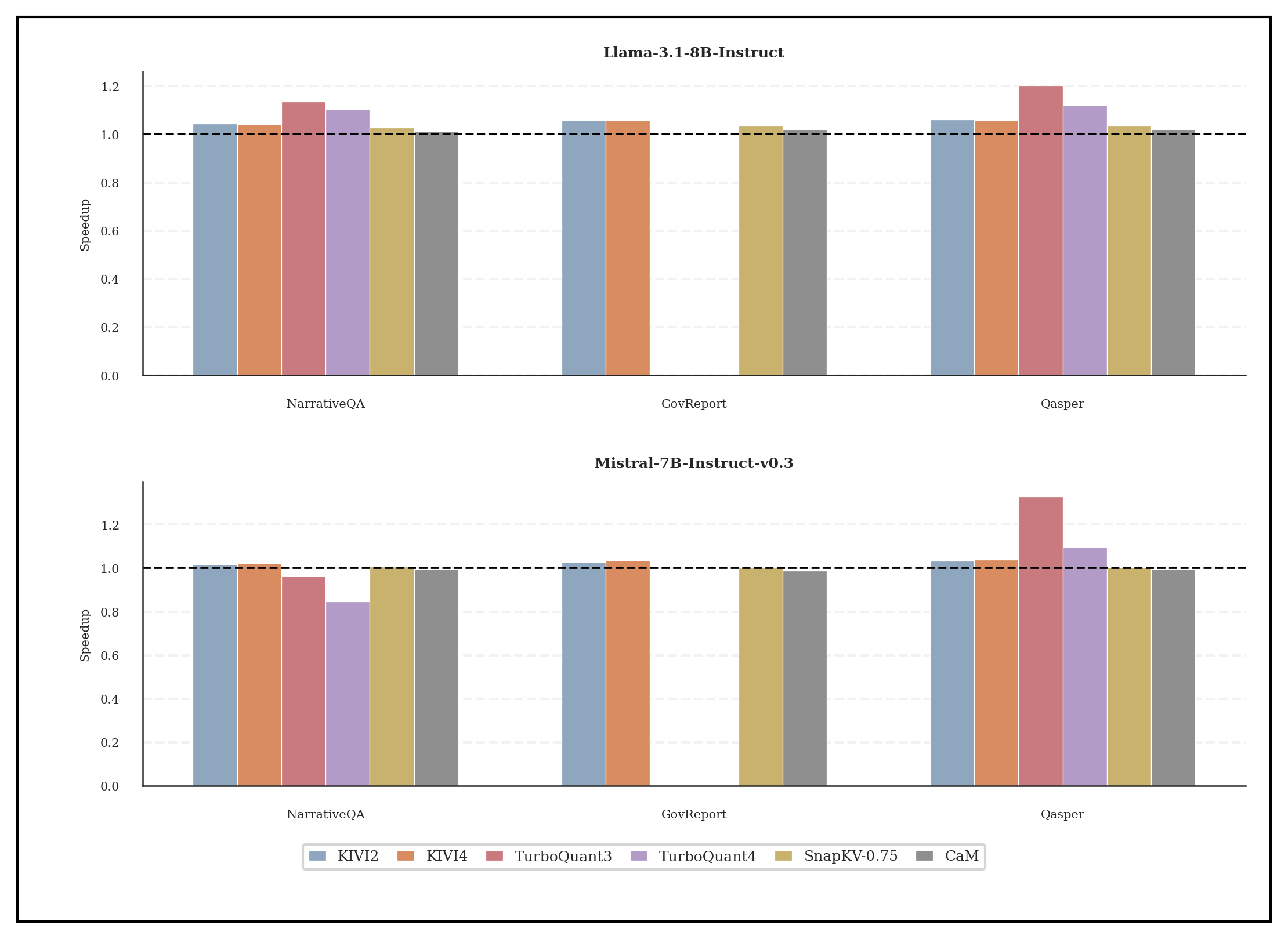}
    \caption{Relative time for TTFT (normalized to All KV) of KV compression methods across models and tasks. \\
    \textit{Note:} We do not present results for Gov\_Report using TurboQuant because it takes upto 3 days for one single run.}
    \label{fig:speed_over_allkv}
\end{figure}

\paragraph{Time to first token} Table \ref{ttft} shows TTFT over different context length buckets for all KV cache compression methods. In terms of TTFT, we observe only moderate differences across KV compression techniques, which are largely consistent across both models, as illustrated in Fig. \ref{fig:speed_over_allkv}. Overall, most methods remain close to the All-KV baseline, confirming that KV cache optimizations have limited impact on prefill latency. KIVI and SnapKV show slightly increased TTFT compared to All-KV, reflecting the additional overhead introduced by quantization and attention-based token selection during the prefill phase. CaM performs similarly, with only minor deviations, as its merging operations are applied periodically and do not heavily impact the initial prefill stage.

In contrast, TurboQuant exhibits the highest TTFT overhead among all methods, particularly for longer contexts such as NarrativeQA and Qasper. This is expected, as TurboQuant applies computationally expensive transformations, including random rotations and residual corrections, which increase the cost of KV cache construction before the first token is generated. Additionally, we observe that TTFT increases with context length across all methods, highlighting that prompt processing remains the dominant factor in prefill latency.

Overall, these results indicate that KV cache compression techniques introduce only limited additional latency for the first token, with the exception of TurboQuant, whose more complex quantization pipeline leads to noticeable overhead. This suggests that most compression methods can be safely applied in latency-sensitive applications without significantly affecting perceived responsiveness, aligning with the observation that TTFT is primarily dominated by prompt encoding rather than KV cache operations.

\begin{figure}[H]
    \centering
    \includegraphics[width=\linewidth]{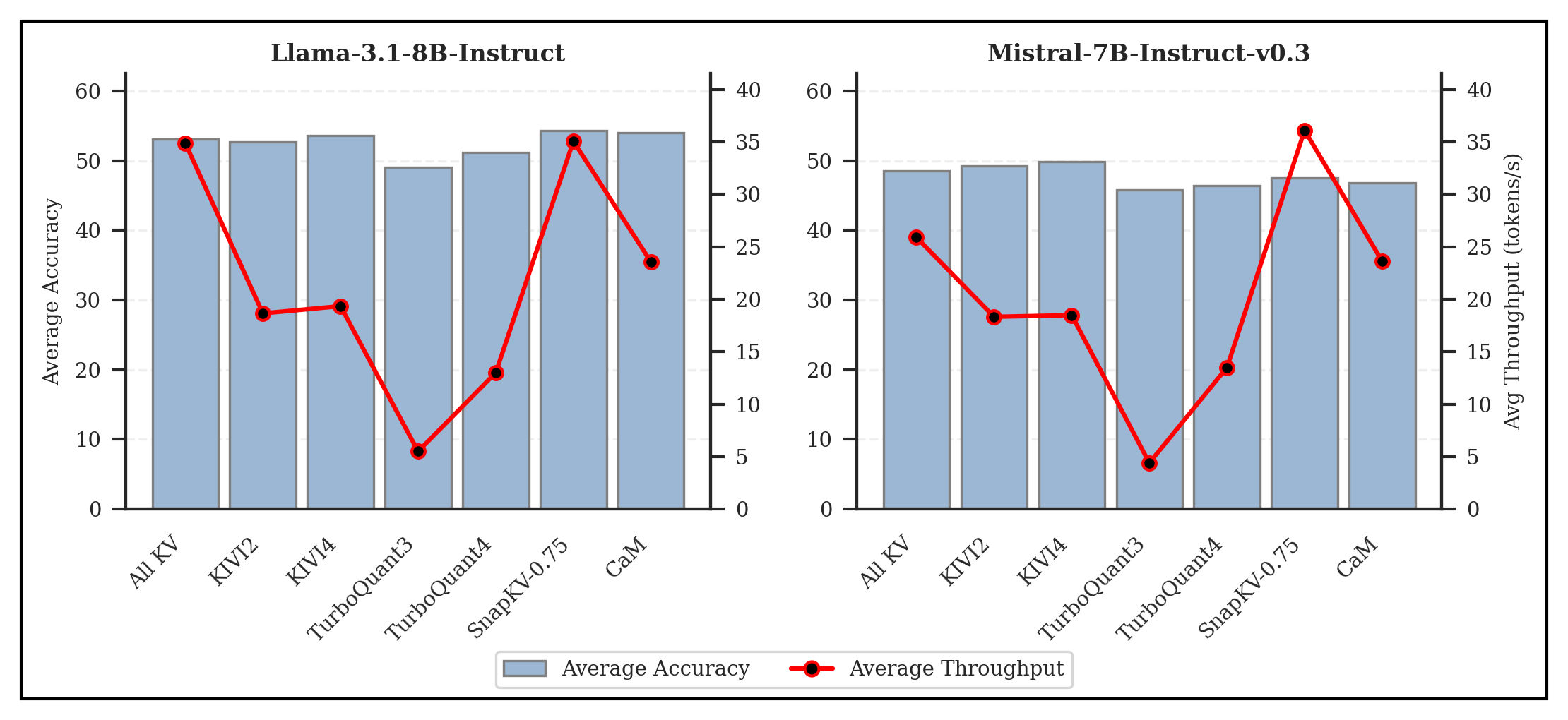}
    \caption{Average accuracy and throughput across different KV compression methods}
    \label{fig:accuracy_throughput_tradeoff}
\end{figure}

\paragraph{Throughput} In terms of throughput, we see large differences between the compression techniques that are consistent across both models as can be seen in Fig. \ref{fig:accuracy_throughput_tradeoff}. SnapKV achieves the best throughput, on par with All-KV on Llama and even surpassing All-KV on Mistral. KIVI and CaM are in the middle, suffering of only modest throughput degradation. The most overhead is introduced in TurboQuant, leading to severe throughput drop.

These results are expected, as SnapKV comes with very low overhead and effectively reduces the \emph{number} of tokens in the KV cache instead of just compressing their representation. CaM also reduces the number of tokens, but induces higher overheads as it redistributes the contributions of evicted tokens. The quantization techniques KIVI and TurboQuant do not reduce the number of tokens in the KV cache, but only compress them. Here, the compression techniques in TurboQuant are more aggressive, leading to higher overheads that have significant impact on throughput.

\paragraph{Compression rate on KV memory} Table~\ref{compression_rates} shows the compression rates of the various KV cache optimizations, broken down across workloads and context-lengths within each workload. This allows for a detailed look into the \emph{effectiveness} of the techniques in achieving their primary goal: reducing the size of the KV cache in GPU memory. We make two major observations, one on stability of compression rates and one on the maximally achievable compression.

\textbf{Stability:} KIVI and SnapKV achieve the most stable and predictable compression rates. There are no major fluctuations across the different workloads and context lengths. TurboQuant is also relatively stable, but shows an outlier with the NarrativeQA workload where the compression rate is slightly higher This can be attributed to its rotation-based quantization scheme, which depends on the statistical distribution of KV vectors; longer and more diverse contexts, such as in NarrativeQA, tend to produce more uniformly distributed representations, enabling slightly more effective compression \cite{dong-etal-2025-longred}. For CaM, we see a completely different picture: Compression lies between extreme compression of 14.2 times (GovReport on 8K+ context length) and \underline{no} compression at all (compression rate of 1.0 for NarrativeQA). This variability stems from CaM’s adaptive, attention-driven merging strategy, which dynamically decides whether tokens should be merged or retained based on their estimated importance. In workloads like GovReport, where redundancy is higher and many tokens can be safely merged, CaM achieves very high compression. However, in tasks such as NarrativeQA, where a larger portion of the context remains relevant for downstream generation, fewer tokens qualify for merging, resulting in minimal compression. From a data management perspective, such unpredcitability of compression effectiveness may make it challenge to operate CaM in real-world deployments. 

\textbf{Effectiveness:} Among the quantization-based compression techniques, KIVI2 shows the highest effectiveness with an average compression rate between 5.16 and 5.32.  The compression rates of TurboQuant3 and Turboquant4 are between those of KIVI2 and KIVI4. SnapKV is also exactly in the middle between both KIVI settings. This shows that, by using the quantization width as a parameter, it is possible to carefully tune the approaches to achieve a desired compression rate. This is good news for practical data management, providing an effective tuning knob that can be adapted to given hardware constraints and workload characteristics. If extreme compression is needed under certain long-context workloads, CaM can be a good choice, as it achieved the highest compression rates under the long-context GovReports tasks (up to 14.2).

\section{Lessons Learned}
\label{sec:lessons}

We present lessons learned and relate them to conventional data management wisdom, comparing our lessons to insights from recent data management papers on related topics where appropriate.

\textbf{KV cache compression can preserve high accuracy.} All-KV was not the consistent winner in terms of accuracy across the various workloads in our benchmark. Instead, under most workloads, one of the compression techniques achieved even better results. This shows that it is not important to retain \emph{all} information in the KV cache, but to identify the \emph{task-relevant} information. Summarizing across all workloads, we show that the decision to use a KV cache compression technique does \underline{not} necessarily mean trading accuracy for memory efficiency. This is different from conventional data management wisdom about lossy compression, where information loss is considered an undesirable but necessary cost to achieve a high compression rate \cite{10.1145/3183713.3196894, 10184656}. 

\textbf{Compression overhead can be amortized by reducing KV cache size.} In terms of throughput, we observe that some compression techniques indeed cause a certain throughput degradation, but not all. By reducing the number of tokens in the KV cache, SnapKV yielded even higher throughput than the All-KV baseline. Again, this is different from conventional data management wisdom, where compression and decompression overheads are considered an additional cost that one has to pay to achieve a reduction in data size. One could compare our results to Isenko et al.~\cite{10.1145/3514221.3517848}, who showed that data compression in machine learning preprocessing pipelines can in some cases be amortized by overcoming communication bottlenecks. 

\textbf{Stable compression rates are preferable in real-world deployments.} While CaM achieves extremely high compression rates on some long-context workloads, it fails to be effective on shorter context workloads. This discrepancy has consequences for practicability. It may be extremely hard to predict under which circumstances an LLM may be used, especially if it accepts prompts from users. Indeed, it is one of the strengths of LLMs that they are general-purpose across a variety of different tasks. In such settings, from a data systems management perspective, having more predictable performance is key. Hence, we recommend using quantization or pruning methods except for scenarios with repeated and uniform tasks, such as using LLM-based summarization for similar kinds of documents.

\textbf{Latency in terms of time-to-first-token is merely affected by KV cache optimizations.} While there are overheads involved in decompressing KV cache (especially under quantization), these do not significantly affect the perceived latency for the user. This is because KV-cache overheads are insignificant when compared to decoding and detokenization phases of LLM inference. As a result, KV cache optimizations can safely be applied in latency-sensitive settings such as user-facing interactions.

\textbf{KV cache optimization trends are similar across LLM models.} We find that the relative performance of the various KV cache compression methods is consistent for both Llama-3.1-8B-Instruct and Mistral-7B-Instruct-v0.3. In particular, KIVI provides consistent accuracy, SnapKV consistently improves throughput, and TurboQuant incurs additional overheads across both models. While absolute performance scores vary due to architectural differences such as sliding-window attention in Mistral, the qualitative trends are consistent. From a data management perspective, this implies that system-level insights from KV cache optimization are transferrable across model families, enabling generalizable optimization strategies without the need for model-specific tuning.

\textbf{Accuracy stability is different in different workloads.} We observe that the impact of KV cache compression on accuracy strongly depends on the task type. Tasks such as few-shot learning are relatively robust to compression, as they rely primarily on recent tokens and local patterns, whereas summarization is highly sensitive to any form of KV modification due to its reliance on global context. Multi-document QA lies in between, benefiting from selective pruning or merging that preserves important information while removing redundancy. From a data management perspective, this implies that workload characteristics such as context redundancy, dependency structure, and sensitivity to global information determine the effectiveness of compression. Consequently, KV cache optimization should be treated as a workload-dependent decision rather than a uniform system configuration.

\textbf{Quantization provides robustness across unknown workloads, but not all methods behave equally.} We observe that KIVI is largely insensitive to workload type, maintaining stable accuracy across QA, few-shot, and summarization tasks. This robustness is due to its asymmetric, data-aware design that takes into account explicitly different outlier structures in keys and values and preserves recent tokens in higher precision with a residual cache. Thus, KIVI preserves both local and global context information even under aggressive compression, making it a reliable default choice when workload characteristics are unknown.

In contrast, TurboQuant is not as robust, despite also being a quantization-based method. Its data-oblivious design relying on random rotations and uniform scalar quantization, leads to small but systematic reconstruction errors on all tokens. While these may be acceptable for local or pattern-based tasks, they accumulate in workloads requiring an understanding of global context, such as summarization, resulting in significant quality degradation. Moreover, TurboQuant introduces higher computational overhead due to its transformation and residual correction steps, which adversely affect system performance. From a data management perspective, this implies that not only the compression paradigm, but also the underlying design principles (data-aware vs. data-oblivious) dictate robustness and suitability for deployment.

\section{Related Work}
\label{sec:related}

Recent work has increasingly framed KV cache optimization as a system-level problem in LLM inference, closely aligned with data management concerns such as memory efficiency, scheduling, and resource utilization. Surveys such as Towards Efficient Large Language Model Serving \cite{xu2026kvcache} emphasize that KV cache optimization spans multiple system dimensions, including execution scheduling, memory placement, and representation design, highlighting that KV cache is a central bottleneck in modern LLM serving systems . Similarly, Li et al. \cite{li2025a} categorize KV cache management techniques into token-level, model-level, and system-level optimizations, reflecting the need for holistic approaches that jointly consider computation and memory trade-offs . Xu et al. \cite{10.1145/3778534.3778567} further show that no single optimization strategy dominates across workloads, and that adaptive, workload-dependent KV cache management is essential for scalable inference .

From a data management perspective, prior work on ML pipelines has demonstrated that performance bottlenecks are often driven by memory movement and data representation rather than pure computation, and that compression can improve system efficiency despite introducing approximation errors~\cite{10.1145/3183713.3196894, 10.1145/3514221.3517848, 10184656}. However, these studies typically assume a fixed trade-off between compression and accuracy, and evaluate optimizations at isolated stages of the pipeline. Earlier works such as CLA~\cite{10.14778/2994509.2994515} considered lossless compression only, which is hard to achieve in KV cache as there is little redundancy in KV cache tensors that could be exploited.

In contrast, our work positions KV cache compression as a core data management problem within LLM inference pipelines, and provides a unified evaluation across accuracy, latency, throughput, and memory under realistic long-context workloads. Unlike prior survey and systems work, which primarily categorize techniques or analyze them in isolation, we empirically show that the effectiveness of KV cache optimization is highly workload-dependent and that compression does not necessarily imply a loss in task quality. This extends existing data management insights by demonstrating that workload-aware KV cache strategies can simultaneously improve both system efficiency and model quality, challenging traditional assumptions about lossy compression in data systems.
\section{Conclusions}
\label{sec:conclusions}

LLM serving pipelines receive growing attention in the data management community, as they involve difficult data management and systems challenges. In this paper, we investigate an important component of LLM serving: the KV cache which stores previously computed attention keys and values so a transformer can reuse them during autoregressive decoding, avoiding recomputation and dramatically speeding up token generation. KV cache optimizations like quantization, pruning, and merging reduce the memory footprint and bandwidth of stored attention keys and values with the goal of enabling faster and more scalable decoding with minimal impact on output quality.

Our benchmarking study reveals various trade-offs in using the existing methods, and provides surprising but practical insights that in parts defy common data management wisdom about compression. Counter-intuitive findings such as the fact that lossy compression could \emph{improve} accuracy in a data system challenge the way we may think about data management for LLMs. At the same time, we found that not all compression techniques are easily deployable in general-purpose LLM serving stacks, such as CaM that could yield by far the best compression rate, or no compression at all, depending on the data set and task. This shows significant challenges when setting up and optimizing an LLM inference system. The data management community with its decade-long experience in query optimization may play a big role in workload-aware optimization of future LLM serving stacks. Our benchmarking study may be a starting point of such explorations.

\section*{Acknowledgments}

We thank the Kuenneth Research Group at the University of Bayreuth for providing access to an NVIDIA A100 40GB GPU, which enabled us to run our benchmarks.


\bibliographystyle{ACM-Reference-Format}
\bibliography{sample}

@String{BIT = "{BIT}" }

@String{Computing = "Computing" }

@String{Computer = "{IEEE} Computer" }

@inproceedings{10.1145/3183713.3196894,
author = {Jiang, Jiawei and Fu, Fangcheng and Yang, Tong and Cui, Bin},
title = {SketchML: Accelerating Distributed Machine Learning with Data Sketches},
year = {2018},
isbn = {9781450347037},
publisher = {Association for Computing Machinery},
address = {New York, NY, USA},
url = {https://doi.org/10.1145/3183713.3196894},
doi = {10.1145/3183713.3196894},
booktitle = {Proceedings of the 2018 International Conference on Management of Data},
pages = {1269–1284},
numpages = {16},
location = {Houston, TX, USA},
series = {SIGMOD '18}
}

@article{10.14778/2994509.2994515,
author = {Elgohary, Ahmed and Boehm, Matthias and Haas, Peter J. and Reiss, Frederick R. and Reinwald, Berthold},
title = {Compressed linear algebra for large-scale machine learning},
year = {2016},
issue_date = {August 2016},
publisher = {VLDB Endowment},
volume = {9},
number = {12},
issn = {2150-8097},
url = {https://doi.org/10.14778/2994509.2994515},
doi = {10.14778/2994509.2994515},
journal = {Proc. VLDB Endow.},
month = aug,
pages = {960–971},
numpages = {12}
}

@article{ge2023model,
  title={Model tells you what to discard: Adaptive kv cache compression for llms},
  author={Ge, Suyu and Zhang, Yunan and Liu, Liyuan and Zhang, Minjia and Han, Jiawei and Gao, Jianfeng},
  journal={arXiv preprint arXiv:2310.01801},
  year={2023}
}

@INPROCEEDINGS{10184656,
  author={Behme, Lennart and Thirumuruganathan, Saravanan and Mahdiraji, Alireza Rezaei and Quiané-Ruiz, Jorge-Arnulfo and Markl, Volker},
  booktitle={2023 IEEE 39th International Conference on Data Engineering (ICDE)}, 
  title={The Art of Losing to Win: Using Lossy Image Compression to Improve Data Loading in Deep Learning Pipelines}, 
  year={2023},
  volume={},
  number={},
  pages={936-949},
  doi={10.1109/ICDE55515.2023.00077}}

@inproceedings{10.1145/3514221.3517848,
author = {Isenko, Alexander and Mayer, Ruben and Jedele, Jeffrey and Jacobsen, Hans-Arno},
title = {Where Is My Training Bottleneck? Hidden Trade-Offs in Deep Learning Preprocessing Pipelines},
year = {2022},
isbn = {9781450392495},
publisher = {Association for Computing Machinery},
address = {New York, NY, USA},
url = {https://doi.org/10.1145/3514221.3517848},
doi = {10.1145/3514221.3517848},
booktitle = {Proceedings of the 2022 International Conference on Management of Data},
pages = {1825–1839},
numpages = {15},
location = {Philadelphia, PA, USA},
series = {SIGMOD '22}
}

@inproceedings{10.5555/3737916.3740097,
author = {He, Yefei and Zhang, Luoming and Wu, Weijia and Liu, Jing and Zhou, Hong and Zhuang, Bohan},
title = {ZipCache: accurate and efficient KV cache quantization with salient token identification},
year = {2024},
isbn = {9798331314385},
publisher = {Curran Associates Inc.},
address = {Red Hook, NY, USA},
booktitle = {Proceedings of the 38th International Conference on Neural Information Processing Systems},
articleno = {2181},
numpages = {21},
location = {Vancouver, BC, Canada},
series = {NIPS '24}
}

@inproceedings{10.5555/3692070.3693381,
author = {Liu, Zirui and Yuan, Jiayi and Jin, Hongye and Zhong, Shaochen (Henry) and Xu, Zhaozhuo and Braverman, Vladimir and Chen, Beidi and Hu, Xia},
title = {KIVI: a tuning-free asymmetric 2bit quantization for KV cache},
year = {2024},
publisher = {JMLR.org},
booktitle = {Proceedings of the 41st International Conference on Machine Learning},
articleno = {1311},
numpages = {13},
location = {Vienna, Austria},
series = {ICML'24}
}

@inproceedings{10.5555/3737916.3737956,
author = {Hooper, Coleman and Kim, Sehoon and Mohammadzadeh, Hiva and Mahoney, Michael W. and Shao, Yakun Sophia and Keutzer, Kurt and Gholami, Amir},
title = {KVQuant: towards 10 million context length LLM inference with KV cache quantization},
year = {2024},
isbn = {9798331314385},
publisher = {Curran Associates Inc.},
address = {Red Hook, NY, USA},
booktitle = {Proceedings of the 38th International Conference on Neural Information Processing Systems},
articleno = {40},
numpages = {34},
location = {Vancouver, BC, Canada},
series = {NIPS '24}
}

@INPROCEEDINGS{11375406,
  author={Cheng, Wen and Dong, Shichen and Qin, Jiayu and Wang, Wei},
  booktitle={2025 IEEE/CVF International Conference on Computer Vision Workshops (ICCVW)}, 
  title={QAQ: Quality Adaptive Quantization for LLM KV Cache}, 
  year={2025},
  volume={},
  number={},
  pages={2563-2571},
  doi={10.1109/ICCVW69036.2025.00267}}

@inproceedings{10.5555/3780338.3781795,
author = {Li, Junyan and Zhang, Yang and Hassan, Muhammad Yusuf and Chafekar, Talha and Cai, Tianle and Ren, Zhile and Guo, Pengsheng and Karimzadeh, Binazir and Reed, Colorado J and Wang, Chong and Gan, Chuang},
title = {CommVQ: commutative vector quantization for KV cache compression},
year = {2025},
publisher = {JMLR.org},
booktitle = {Proceedings of the 42nd International Conference on Machine Learning},
articleno = {1457},
numpages = {15},
location = {Vancouver, Canada},
series = {ICML'25}
}

@inproceedings{
wu2026polarquant,
title={PolarQuant: Leveraging Polar Transformation for Key Cache Quantization and Decoding Acceleration},
author={Songhao Wu and Ang Lv and xiao feng and Yufei zhang and Xun Zhang and Guojun Yin and Wei Lin and Rui Yan},
booktitle={The Thirty-ninth Annual Conference on Neural Information Processing Systems},
year={2026},
url={https://openreview.net/forum?id=JCTTLKEBza}
}

@inproceedings{
zandieh2026turboquant,
title={TurboQuant: Online Vector Quantization with Near-optimal Distortion Rate},
author={Amir Zandieh and Majid Daliri and Majid Hadian and Vahab Mirrokni},
booktitle={The Fourteenth International Conference on Learning Representations},
year={2026},
url={https://openreview.net/forum?id=tO3ASKZlok}
}

@inproceedings{10.5555/3737916.3738638,
author = {Li, Yuhong and Huang, Yingbing and Yang, Bowen and Venkitesh, Bharat and Locatelli, Acyr and Ye, Hanchen and Cai, Tianle and Lewis, Patrick and Chen, Deming},
title = {SnapKV: LLM knows what you are looking for before generation},
year = {2024},
isbn = {9798331314385},
publisher = {Curran Associates Inc.},
address = {Red Hook, NY, USA},
booktitle = {Proceedings of the 38th International Conference on Neural Information Processing Systems},
articleno = {722},
numpages = {24},
location = {Vancouver, BC, Canada},
series = {NIPS '24}
}

@inproceedings{10.5555/3666122.3667628,
author = {Zhang, Zhenyu and Sheng, Ying and Zhou, Tianyi and Chen, Tianlong and Zheng, Lianmin and Cai, Ruisi and Song, Zhao and Tian, Yuandong and R\'{e}, Christopher and Barrett, Clark and Wang, Zhangyang and Chen, Beidi},
title = {H2O: heavy-hitter oracle for efficient generative inference of large language models},
year = {2023},
publisher = {Curran Associates Inc.},
address = {Red Hook, NY, USA},
booktitle = {Proceedings of the 37th International Conference on Neural Information Processing Systems},
articleno = {1506},
numpages = {50},
location = {New Orleans, LA, USA},
series = {NIPS '23}
}

@inproceedings{
xiao2024efficient,
title={Efficient Streaming Language Models with Attention Sinks},
author={Guangxuan Xiao and Yuandong Tian and Beidi Chen and Song Han and Mike Lewis},
booktitle={The Twelfth International Conference on Learning Representations},
year={2024},
url={https://openreview.net/forum?id=NG7sS51zVF}
}

@inproceedings{devoto-etal-2024-simple,
    title = "A Simple and Effective $L\_2$ Norm-Based Strategy for {KV} Cache Compression",
    author = "Devoto, Alessio  and
      Zhao, Yu  and
      Scardapane, Simone  and
      Minervini, Pasquale",
    editor = "Al-Onaizan, Yaser  and
      Bansal, Mohit  and
      Chen, Yun-Nung",
    booktitle = "Proceedings of the 2024 Conference on Empirical Methods in Natural Language Processing",
    month = nov,
    year = "2024",
    address = "Miami, Florida, USA",
    publisher = "Association for Computational Linguistics",
    url = "https://aclanthology.org/2024.emnlp-main.1027/",
    doi = "10.18653/v1/2024.emnlp-main.1027",
    pages = "18476--18499"
}

@inproceedings{chitty-venkata-etal-2026-pagedeviction,
    title = "{P}aged{E}viction: Structured Block-wise {KV} Cache Pruning for Efficient Large Language Model Inference",
    author = "Chitty-Venkata, Krishna Teja  and
      Ye, Jie  and
      Raskar, Siddhisanket  and
      Kougkas, Anthony  and
      Sun, Xian  and
      Emani, Murali  and
      Vishwanath, Venkatram  and
      Nicolae, Bogdan",
    editor = "Demberg, Vera  and
      Inui, Kentaro  and
      Marquez, Llu{\'i}s",
    booktitle = "Findings of the {A}ssociation for {C}omputational {L}inguistics: {EACL} 2026",
    month = mar,
    year = "2026",
    address = "Rabat, Morocco",
    publisher = "Association for Computational Linguistics",
    url = "https://aclanthology.org/2026.findings-eacl.168/",
    doi = "10.18653/v1/2026.findings-eacl.168",
    pages = "3207--3218",
    ISBN = "979-8-89176-386-9"
}

@InProceedings{pmlr-v235-zhang24n,
  title = 	 {{C}a{M}: Cache Merging for Memory-efficient {LLM}s Inference},
  author =       {Zhang, Yuxin and Du, Yuxuan and Luo, Gen and Zhong, Yunshan and Zhang, Zhenyu and Liu, Shiwei and Ji, Rongrong},
  booktitle = 	 {Proceedings of the 41st International Conference on Machine Learning},
  pages = 	 {58840--58850},
  year = 	 {2024},
  editor = 	 {Salakhutdinov, Ruslan and Kolter, Zico and Heller, Katherine and Weller, Adrian and Oliver, Nuria and Scarlett, Jonathan and Berkenkamp, Felix},
  volume = 	 {235},
  series = 	 {Proceedings of Machine Learning Research},
  month = 	 {21--27 Jul},
  publisher =    {PMLR},
  url = 	 {https://proceedings.mlr.press/v235/zhang24n.html}
}

@article{song2024hierarchical,
  title={Hierarchical context merging: Better long context understanding for pre-trained LLMs},
  author={Song, Woomin and Oh, Seunghyuk and Mo, Sangwoo and Kim, Jaehyung and Yun, Sukmin and Ha, Jung-Woo and Shin, Jinwoo},
  journal={ICLR},
  year={2024}
}

@inproceedings{bolya2022tome,
  title={Token Merging: Your {ViT} but Faster},
  author={Bolya, Daniel and Fu, Cheng-Yang and Dai, Xiaoliang and Zhang, Peizhao and Feichtenhofer, Christoph and Hoffman, Judy},
  booktitle={International Conference on Learning Representations},
  year={2023}
}

@misc{munkhdalai2024leavecontextbehindefficient,
      title={Leave No Context Behind: Efficient Infinite Context Transformers with Infini-attention}, 
      author={Tsendsuren Munkhdalai and Manaal Faruqui and Siddharth Gopal},
      year={2024},
      eprint={2404.07143},
      archivePrefix={arXiv},
      primaryClass={cs.CL},
      url={https://arxiv.org/abs/2404.07143}, 
}

@inproceedings{
kim2024compressed,
title={Compressed Context Memory for Online Language Model Interaction},
author={Jang-Hyun Kim and Junyoung Yeom and Sangdoo Yun and Hyun Oh Song},
booktitle={The Twelfth International Conference on Learning Representations},
year={2024},
url={https://openreview.net/forum?id=64kSvC4iPg}
}

@inproceedings{zuhri-etal-2025-mlkv,
    title = "{MLKV}: Multi-Layer Key-Value Heads for Memory Efficient Transformer Decoding",
    author = "Zuhri, Zayd Muhammad Kawakibi  and
      Adilazuarda, Muhammad Farid  and
      Purwarianti, Ayu  and
      Aji, Alham Fikri",
    editor = "Chiruzzo, Luis  and
      Ritter, Alan  and
      Wang, Lu",
    booktitle = "Findings of the Association for Computational Linguistics: NAACL 2025",
    month = apr,
    year = "2025",
    address = "Albuquerque, New Mexico",
    publisher = "Association for Computational Linguistics",
    url = "https://aclanthology.org/2025.findings-naacl.305/",
    doi = "10.18653/v1/2025.findings-naacl.305",
    pages = "5531--5540",
    ISBN = "979-8-89176-195-7"
}

@inproceedings{
sun2024you,
title={You Only Cache Once: Decoder-Decoder Architectures for Language Models},
author={Yutao Sun and Li Dong and Yi Zhu and Shaohan Huang and Wenhui Wang and Shuming Ma and Quanlu Zhang and Jianyong Wang and Furu Wei},
booktitle={The Thirty-eighth Annual Conference on Neural Information Processing Systems},
year={2024},
url={https://openreview.net/forum?id=25Ioxw576r}
}

@inproceedings{10.5555/3737916.3742359,
author = {Liu, Akide and Liu, Jing and Pan, Zizheng and He, Yefei and Haffari, Gholamreza and Zhuang, Bohan},
title = {MiniCache: KV cache compression in depth dimension for large language models},
year = {2024},
isbn = {9798331314385},
publisher = {Curran Associates Inc.},
address = {Red Hook, NY, USA},
booktitle = {Proceedings of the 38th International Conference on Neural Information Processing Systems},
articleno = {4443},
numpages = {35},
location = {Vancouver, BC, Canada},
series = {NIPS '24}
}

@misc{
yang2024kvsharer,
title={{KVS}harer: Efficient Inference via Layer-Wise Dissimilar {KV} Cache Sharing},
author={Yifei Yang and zouying cao and Qiguang Chen and Libo Qin and Dongjie Yang and Zhi Chen and hai zhao},
year={2024},
url={https://openreview.net/forum?id=2Akf4BBCKo}
}

@inproceedings{
ainslie2023gqa,
title={{GQA}: Training Generalized Multi-Query Transformer Models from Multi-Head Checkpoints},
author={Joshua Ainslie and James Lee-Thorp and Michiel de Jong and Yury Zemlyanskiy and Federico Lebron and Sumit Sanghai},
booktitle={The 2023 Conference on Empirical Methods in Natural Language Processing},
year={2023},
url={https://openreview.net/forum?id=hmOwOZWzYE}
}

@misc{shazeer2019fasttransformerdecodingwritehead,
      title={Fast Transformer Decoding: One Write-Head is All You Need}, 
      author={Noam Shazeer},
      year={2019},
      eprint={1911.02150},
      archivePrefix={arXiv},
      primaryClass={cs.NE},
      url={https://arxiv.org/abs/1911.02150}, 
}

@inproceedings{
chang2025palu,
title={Palu: {KV}-Cache Compression with Low-Rank Projection},
author={Chi-Chih Chang and Wei-Cheng Lin and Chien-Yu Lin and Chong-Yan Chen and Yu-Fang Hu and Pei-Shuo Wang and Ning-Chi Huang and Luis Ceze and Mohamed S. Abdelfattah and Kai-Chiang Wu},
booktitle={The Thirteenth International Conference on Learning Representations},
year={2025},
url={https://openreview.net/forum?id=LWMS4pk2vK}
}

@misc{
zhang2025lorc,
title={Lo{RC}: Low-Rank Compression for {LLM}s {KV} Cache with a Progressive Compression Strategy},
author={Rongzhi Zhang and Kuan Wang and Liyuan Liu and Shuohang Wang and Hao Cheng and Chao Zhang and yelong shen},
year={2025},
url={https://openreview.net/forum?id=NI8AUSAc4i}
}

@inproceedings{bai2024longbench,
    title = "{L}ong{B}ench: A Bilingual, Multitask Benchmark for Long Context Understanding",
    author = "Bai, Yushi and Lv, Xin  and Zhang, Jiajie  and Lyu, Hongchang  and
      Tang, Jiankai  and Huang, Zhidian  and Du, Zhengxiao  and Liu, Xiao  and Zeng, Aohan  and Hou, Lei  and Dong, Yuxiao  and Tang, Jie  and Li, Juanzi",
    booktitle = "Proceedings of the 62nd Annual Meeting of the Association for Computational Linguistics (Volume 1: Long Papers)",
    month = aug,
    year = "2024",
    address = "Bangkok, Thailand",
    publisher = "Association for Computational Linguistics",
    url = "https://aclanthology.org/2024.acl-long.172",
    doi = "10.18653/v1/2024.acl-long.172",
    pages = "3119--3137",
}

@misc{grattafiori2024llama3herdmodels,
      title={The Llama 3 Herd of Models}, 
      author={Aaron Grattafiori and Abhimanyu Dubey et al.},
      year={2024},
      eprint={2407.21783},
      archivePrefix={arXiv},
      primaryClass={cs.AI},
      url={https://arxiv.org/abs/2407.21783}, 
}

@misc{jiang2023mistral7b,
      title={Mistral 7B}, 
      author={Albert Q. Jiang and Alexandre Sablayrolles and Arthur Mensch and Chris Bamford and Devendra Singh Chaplot and Diego de las Casas and Florian Bressand and Gianna Lengyel and Guillaume Lample and Lucile Saulnier and Lélio Renard Lavaud and Marie-Anne Lachaux and Pierre Stock and Teven Le Scao and Thibaut Lavril and Thomas Wang and Timothée Lacroix and William El Sayed},
      year={2023},
      eprint={2310.06825},
      archivePrefix={arXiv},
      primaryClass={cs.CL},
      url={https://arxiv.org/abs/2310.06825}, 
}

@inproceedings{yuan-etal-2024-kv,
    title = "{KV} Cache Compression, But What Must We Give in Return? A Comprehensive Benchmark of Long Context Capable Approaches",
    author = "Yuan, Jiayi  and
      Liu, Hongyi  and
      Zhong, Shaochen  and
      Chuang, Yu-Neng  and
      Li, Songchen  and
      Wang, Guanchu  and
      Le, Duy  and
      Jin, Hongye  and
      Chaudhary, Vipin  and
      Xu, Zhaozhuo  and
      Liu, Zirui  and
      Hu, Xia",
    editor = "Al-Onaizan, Yaser  and
      Bansal, Mohit  and
      Chen, Yun-Nung",
    booktitle = "Findings of the Association for Computational Linguistics: EMNLP 2024",
    month = nov,
    year = "2024",
    address = "Miami, Florida, USA",
    publisher = "Association for Computational Linguistics",
    url = "https://aclanthology.org/2024.findings-emnlp.266/",
    doi = "10.18653/v1/2024.findings-emnlp.266",
    pages = "4623--4648"
}

@inproceedings{10.1145/3778534.3778567,
author = {Liu, Yanyu and Fu, Jingying and Liu, Sixiang and Zou, Yitian and Zhang, Shouhua and Zhou, Jiehan},
title = {KV Cache Compression for Inference Efficiency in LLMs: A Review},
year = {2026},
isbn = {9798400719356},
publisher = {Association for Computing Machinery},
address = {New York, NY, USA},
url = {https://doi.org/10.1145/3778534.3778567},
doi = {10.1145/3778534.3778567},
booktitle = {Proceedings of the 4th International Conference on Artificial Intelligence and Intelligent Information Processing},
pages = {207–212},
numpages = {6},
location = {
},
series = {AIIIP '25}
}

@inproceedings{duan-etal-2024-botchat,
    title = "{B}ot{C}hat: Evaluating {LLM}s' Capabilities of Having Multi-Turn Dialogues",
    author = "Duan, Haodong  and
      Wei, Jueqi  and
      Wang, Chonghua  and
      Liu, Hongwei  and
      Fang, Yixiao  and
      Zhang, Songyang  and
      Lin, Dahua  and
      Chen, Kai",
    editor = "Duh, Kevin  and
      Gomez, Helena  and
      Bethard, Steven",
    booktitle = "Findings of the Association for Computational Linguistics: NAACL 2024",
    month = jun,
    year = "2024",
    address = "Mexico City, Mexico",
    publisher = "Association for Computational Linguistics",
    url = "https://aclanthology.org/2024.findings-naacl.201/",
    doi = "10.18653/v1/2024.findings-naacl.201",
    pages = "3184--3200"
}

@inproceedings{10.1145/3663741.3664785,
author = {Barbon Junior, Sylvio and Ceravolo, Paolo and Groppe, Sven and Jarrar, Mustafa and Maghool, Samira and S\`{e}des, Florence and Sahri, Soror and Van Keulen, Maurice},
title = {Are Large Language Models the New Interface for Data Pipelines?},
year = {2024},
isbn = {9798400706790},
publisher = {Association for Computing Machinery},
address = {New York, NY, USA},
url = {https://doi.org/10.1145/3663741.3664785},
doi = {10.1145/3663741.3664785},
booktitle = {Proceedings of the International Workshop on Big Data in Emergent Distributed Environments},
articleno = {6},
numpages = {6},
location = {Santiago, AA, Chile},
series = {BiDEDE '24}
}

@article{10.14778/3750601.3750703,
author = {Pan, James and Li, Guoliang},
title = {Database Perspective on LLM Inference Systems},
year = {2025},
issue_date = {August 2025},
publisher = {VLDB Endowment},
volume = {18},
number = {12},
issn = {2150-8097},
url = {https://doi.org/10.14778/3750601.3750703},
doi = {10.14778/3750601.3750703},
journal = {Proc. VLDB Endow.},
month = aug,
pages = {5504–5507},
numpages = {4}
}

@article{10.1145/3749168,
author = {Li, Yuhang and Gu, Rong and Huan, Chengying and Wang, Zhibin and Yao, Renjie and Tian, Chen and Chen, Guihai},
title = {HotPrefix: Hotness-Aware KV Cache Scheduling for Efficient Prefix Sharing in LLM Inference Systems},
year = {2025},
issue_date = {September 2025},
publisher = {Association for Computing Machinery},
address = {New York, NY, USA},
volume = {3},
number = {4},
url = {https://doi.org/10.1145/3749168},
doi = {10.1145/3749168},
journal = {Proc. ACM Manag. Data},
month = sep,
articleno = {250},
numpages = {27}
}

@inproceedings{10.5555/3768039.3768067,
author = {Wang, Jiahao and Han, Jinbo and Wei, Xingda and Shen, Sijie and Zhang, Dingyan and Fang, Chenguang and Chen, Rong and Yu, Wenyuan and Chen, Haibo},
title = {KVCache cache in the wild: characterizing and optimizing KVCache cache at a large cloud provider},
year = {2025},
isbn = {978-1-939133-48-9},
publisher = {USENIX Association},
address = {USA},
booktitle = {Proceedings of the 2025 USENIX Conference on Usenix Annual Technical Conference},
articleno = {28},
numpages = {18},
location = {Boston, MA, USA},
series = {USENIX ATC '25}
}

@article{10.1145/3769778,
author = {Yuan, Hao and Ai, Xin and Wang, Qiange and Li, Peizheng and Yu, Jiayang and Chen, Chaoyi and Yang, Xinbo and Zhang, Yanfeng and Fu, Zhenbo and Wen, Yingyou and Yu, Ge},
title = {DepCache: A KV Cache Management Framework for GraphRAG with Dependency Attention},
year = {2025},
issue_date = {December 2025},
publisher = {Association for Computing Machinery},
address = {New York, NY, USA},
volume = {3},
number = {6},
url = {https://doi.org/10.1145/3769778},
doi = {10.1145/3769778},
journal = {Proc. ACM Manag. Data},
month = dec,
articleno = {313},
numpages = {29}
}

@article{10.14778/3685800.3685838,
author = {Li, Guoliang and Zhou, Xuanhe and Zhao, Xinyang},
title = {LLM for Data Management},
year = {2024},
issue_date = {August 2024},
publisher = {VLDB Endowment},
volume = {17},
number = {12},
issn = {2150-8097},
url = {https://doi.org/10.14778/3685800.3685838},
doi = {10.14778/3685800.3685838},
journal = {Proc. VLDB Endow.},
month = aug,
pages = {4213–4216},
numpages = {4}
}

@inproceedings{liu2024cachegen,
  title={Cachegen: Kv cache compression and streaming for fast large language model serving},
  author={Liu, Yuhan and Li, Hanchen and Cheng, Yihua and Ray, Siddhant and Huang, Yuyang and Zhang, Qizheng and Du, Kuntai and Yao, Jiayi and Lu, Shan and Ananthanarayanan, Ganesh and others},
  booktitle={Proceedings of the ACM SIGCOMM 2024 Conference},
  pages={38--56},
  year={2024}
}

@inproceedings{
dao2024flashattention,
title={FlashAttention-2: Faster Attention with Better Parallelism and Work Partitioning},
author={Tri Dao},
booktitle={The Twelfth International Conference on Learning Representations},
year={2024},
url={https://openreview.net/forum?id=mZn2Xyh9Ec}
}

@misc{
devoto2026expected,
title={Expected Attention: {KV} Cache Compression by Estimating Attention From Future Queries Distribution},
author={Alessio Devoto and Maximilian Jeblick and Simon J{\'e}gou},
year={2026},
url={https://openreview.net/forum?id=VmojW15eRc}
}

@misc{hory2026turboquant,
  author       = {Omar Hory},
  title        = {TurboQuant: Open-source implementation of Google's TurboQuant},
  year         = {2026},
  howpublished = {\url{https://github.com/OmarHory/turboquant}},
  note         = {Accessed: 2026-04-29}
}

@inproceedings{
wang2025llms,
title={{LLM}s Know What to Drop: Self-Attention Guided {KV} Cache Eviction for Efficient Long-Context Inference},
author={Guangtao Wang and Shubhangi Upasani and Chen Wu and Darshan Gandhi and Jonathan Lingjie Li and Changran Hu and Bo Li and Urmish Thakker},
booktitle={Sparsity in LLMs (SLLM): Deep Dive into Mixture of Experts, Quantization, Hardware, and Inference},
year={2025},
url={https://openreview.net/forum?id=qg9dlCcNzr}
}

@inproceedings{dong-etal-2025-longred,
    title = "{L}ong{R}e{D}: Mitigating Short-Text Degradation of Long-Context Large Language Models via Restoration Distillation",
    author = "Dong, Zican  and
      Li, Junyi  and
      Jiang, Jinhao  and
      Xu, Mingyu  and
      Zhao, Wayne Xin  and
      Wang, Bingning  and
      Chen, Weipeng",
    editor = "Che, Wanxiang  and
      Nabende, Joyce  and
      Shutova, Ekaterina  and
      Pilehvar, Mohammad Taher",
    booktitle = "Proceedings of the 63rd Annual Meeting of the Association for Computational Linguistics (Volume 1: Long Papers)",
    month = jul,
    year = "2025",
    address = "Vienna, Austria",
    publisher = "Association for Computational Linguistics",
    url = "https://aclanthology.org/2025.acl-long.524/",
    doi = "10.18653/v1/2025.acl-long.524",
    pages = "10687--10707",
    ISBN = "979-8-89176-251-0"
}

@article{xu2026kvcache,
  title={Towards Efficient Large Language Model Serving:
A Survey on System-Aware KV Cache Optimization},
  author={Jiantong Jiang and Peiyu Yang and Rui Zhang and  Feng Liu},
  journal={TechRxiv},
  year={2026},
  doi={10.36227/techrxiv.176046306.66521015/v3}
}

@article{
li2025a,
title={A Survey on Large Language Model Acceleration based on {KV} Cache Management},
author={Haoyang LI and Yiming Li and Anxin Tian and Tianhao Tang and Zhanchao Xu and Xuejia Chen and Nicole HU and Wei Dong and Li Qing and Lei Chen},
journal={Transactions on Machine Learning Research},
issn={2835-8856},
year={2025},
url={https://openreview.net/forum?id=z3JZzu9EA3},
note={}
}

\end{document}